\documentclass[12pt]{article}
\usepackage{amsmath,amsfonts}
\usepackage{graphicx}
\usepackage[T1]{fontenc}
\usepackage{comment,pdflscape}
\usepackage{multirow}
\usepackage{lmodern}
\usepackage{subcaption}
\usepackage{wrapfig}
\usepackage{enumerate} 
\usepackage{array}
\usepackage{booktabs}
\usepackage{tabularx, booktabs, ragged2e}
\usepackage{rotating} 
\usepackage{adjustbox}
\usepackage{tikz}
\usetikzlibrary{shapes, arrows.meta, positioning, fit, backgrounds,calc}
\usepackage{tabularray}
\PassOptionsToPackage{export}{adjustbox}
\usepackage{helvet}
\definecolor{primary}{HTML}{6A9FB5}     
\definecolor{secondary}{HTML}{A2B9BC}   
\definecolor{accent}{HTML}{C7B8A4}      
\definecolor{highlight}{HTML}{F2D7B5}   
\definecolor{textcol}{HTML}{2C3E50}     
\usepackage{hyperref}
\usepackage{geometry}
\usepackage{tabularx}
\usepackage{float,xcolor}
\usepackage{caption}
\newcommand{\model}{{\normalfont\textsc{TwinFormer}}}
\title{\model: A Dual-Level Transformer for Long-Sequence Time-Series Forecasting}
\author{
 Mahima Kumavat\thanks{Email: mahimak@iimidr.ac.in}~ and 
    Aditya Maheshwari\thanks{Email: adityam@iimidr.ac.in} \\
   {\small  Operations Management and Quantitative Techniques Area, }\\{ \small  Indian Institute of Management Indore}
}
\date{}
\begin{document}
\maketitle
\begin{abstract}


\model\ is a hierarchical Transformer for long-sequence time-series forecasting. 
It divides the input into non-overlapping temporal patches and processes them in two stages: 
(1) a Local Informer with top-$k$ Sparse Attention models intra-patch dynamics, followed by mean pooling; 
(2) a Global Informer captures long-range inter-patch dependencies using the same top-$k$ attention. 
A lightweight GRU aggregates the globally contextualized patch tokens for direct multi-horizon prediction. 
The resulting architecture achieves linear $O(kLd)$ time and memory complexity. On eight real-world benchmarking datasets from six different domains including weather, stock price, temperature, power consumption, electricity, and disease, and forecasting horizons 96-720, \model\ secures 27 positions in top two out of 34. Out of the 27, it achieves best performance on MAE and RMSE at 17 places and $10$ at second best place on MAE and RMSE. This consistently outperforms PatchTST, iTransformer, FEDformer, Informer, and vanilla Transformers. Ablations confirm the superiority of top-$k$ Sparse Attention over ProbSparse and the effectiveness of GRU-based aggregation. Code is
available at this repository: \url{https://github.com/Mahimakumavat1205/TwinFormer}.

\end{abstract}
\section{Introduction}

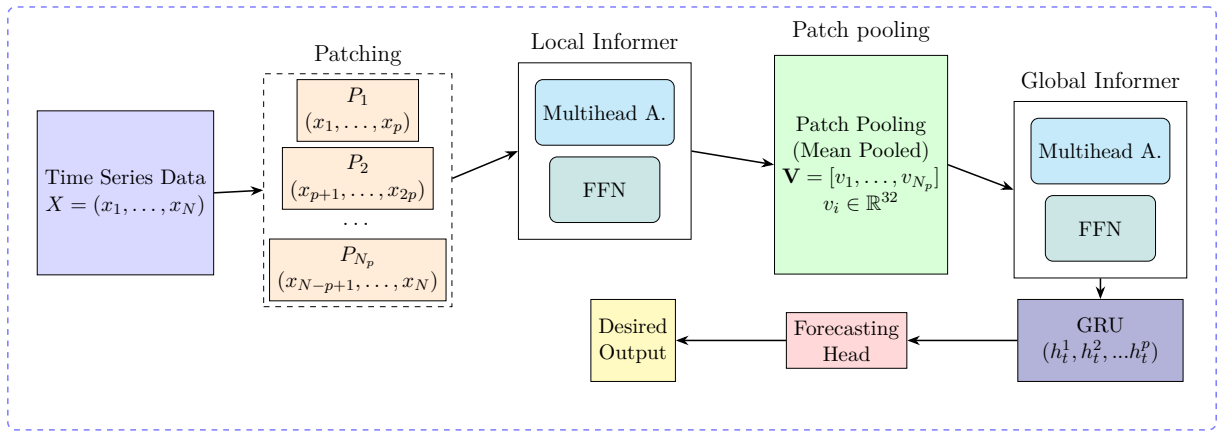
\begin{figure}[ht!]
\centering
\begin{adjustbox}{max width=\textwidth, max height=\textheight, keepaspectratio}

\begin{tikzpicture}[
    data/.style={draw, rectangle, fill=blue!15, minimum height=3cm, minimum width=0.8cm, align=center, font=\small},
    patch_box/.style={draw, rectangle, fill=orange! 15, minimum height=0.5cm, minimum width=1.5cm, align=center, font=\small},
    informer_base/.style={draw, rounded corners, minimum height=1.5cm, minimum width=2.5cm, align=center, font=\small},
    attn_block/.style={informer_base, fill=cyan! 20, minimum height=1.2cm, minimum width=2cm},
    ffn_block/.style={informer_base, fill=teal! 20, minimum height=1.2cm, minimum width=2cm},
    pooling/.style={draw, rectangle, fill=green! 15, minimum height=4cm, minimum width=2.5cm, align=center, font=\small},
    rnn/.style={draw, rectangle, fill=blue!50!black! 30, minimum height=1.5cm, minimum width=3cm, align=center, font=\small},
    linear/.style={draw, rectangle, fill=pink! 60, minimum height=1cm, minimum width=1.5cm, align=center, font=\small},
    output_node/.style={draw, rectangle, fill=yellow! 30, minimum height=1.5cm, minimum width=0.1cm, align=center, font=\small},
    main_flow/.style={draw, ->, thick, >=Stealth},
    minor_flow/.style={draw, ->, thin, dashed},
    vertical_link/.style={draw, ->, thick, blue!70},
    frame_group/.style={draw, rounded corners, thick, blue!50, dashed, minimum height=5cm, minimum width=7cm}
]

\node (TS) [data] at (0,0) {Time Series Data\\ $X = (x_1, \dots, x_N)$};

\node (P_top) [patch_box, right=1.5cm of TS.north east, anchor=west] {$P_1$\\ $(x_1, \dots, x_p)$};
\node (P_mid) [patch_box, below=0.1cm of P_top] {$P_2$\\ $(x_{p+1}, \dots, x_{2p})$};
\node (P_dots) [align=center, below=0.1cm of P_mid] {$\dots$};
\node (P_bot) [patch_box, below=0.1cm of P_dots] {$P_{N_p}$\\ $(x_{N-p+1}, \dots, x_{N})$};
\node [above=0.1cm of P_top, align=center] {Patching};

\node (Patch_Env) [draw, dashed, fit=(P_top) (P_bot), inner sep=0.1cm] {};
\draw [main_flow] (TS.east) -- (Patch_Env.west);

\node (LIF) [attn_block, right=1.5cm of Patch_Env.east|-P_top.north east, anchor=north west] {Multihead A.};
\node (FFN) [ffn_block, below=0.2cm of LIF] {FFN};
\node (LocalBlock) [draw, fit=(LIF) (FFN), inner sep=0.3cm] {};
\draw [main_flow] (Patch_Env.east|-P_mid.center) -- (LocalBlock.west);
\node [align=center] at (LocalBlock.north) [above=0.1cm] {Local Informer};

\node (Pool) [pooling, right=1.5cm of LocalBlock, yshift=-0.25cm] {Patch Pooling\\ (Mean Pooled)\\ $\mathbf{V} = [v_1, \dots, v_{N_p}]$ \\ $v_i \in \mathbb{R}^{32}$};
\node [above=0.1cm of Pool] {Patch pooling};
\draw [main_flow] (LocalBlock.east) -- (Pool.west);

\node (GIF) [attn_block, right=1.5cm of Pool, yshift=0.25cm] {Multihead A.};
\node (FFN2) [ffn_block, below=0.2cm of GIF] {FFN};
\node (GlobalBlock) [draw, fit=(GIF) (FFN2), inner sep=0.3cm] {};
\draw [main_flow] (Pool.east) -- (GlobalBlock.west);
\node [align=center] at (GlobalBlock.north) [above=0.1cm] {Global Informer};

\node (GRU) [rnn, below=2cm of GlobalBlock.center] {GRU\\ ($h_t^1, h_t^2,...h_t^p $)}; 
\draw [main_flow] (GlobalBlock.south) -- (GRU.north);

\node (Linear) [linear, left=2cm of GRU] {Forecasting \\ Head};
\draw [main_flow] (GRU.west) -- (Linear.east);

\node (Desired) [output_node, left=2cm of Linear] {Desired\\ Output};
\draw [main_flow] (Linear.west) -- (Desired.east);

\node (Frame) [frame_group, fit = (TS) (Desired) (GRU) (LIF) (FFN) (Pool) (Patch_Env) (GIF) (GlobalBlock), inner xsep=15pt, inner ysep=25pt, minimum height=5cm, minimum width=7cm] {}; 

\node [above=0.9cm of Frame.north, font=\bfseries\Large, text=blue!70!black] {};

\end{tikzpicture}

\end{adjustbox}

\caption{Architecture of Proposed model}
\label{fig:architechture}
\end{figure}

Long Sequence Time Series Forecasting (LSTSF) is a big challenge in numerous real-world domains, including energy management, hhealthcare monitoring, financial modelling and climate science, where input sequences routinely exceed $10^4$--$10^5$ time steps and accurate multi-horizon predictions are required. Classical statistical methods and early recurrent architectures (e.g., LSTM, GRU) struggle with vanishing gradients and inherently sequential computation, while vanilla Transformers \cite{vaswani2017attention} incur prohibitive $O(L^2)$ complexity, rendering them impractical for very long sequences. However, the quadratic computational complexity $O(L^2)$ of vanilla Transformers with respect to sequence length severely limits their applicability for LSTSF \cite{child2019sparse}. Sparse and low-rank Transformers (Informer \cite{informer2021}, Autoformer \cite{wu2021autoformer}, FEDformer \cite{zhou2022fedformer}) and patch-based tokenisation strategies (PatchTST \cite{nie2023patchtst}, iTransformer \cite{Liu2023-sd}) now dominate leaderboards by reducing sequence length and attention complexity. Despite these advancements, these models still process time series data at a flat sequence, neglecting the hierarchical nature of real-world temporal data and struggling with long-range dependencies and computational inefficiency as sequence length increases. \textsc{TimesNet} \cite{wu2023timesnet} models time series at multiple periodicities using $2D$ temporal maps, whereas \textsc{Hformer} \cite{liu2022time} explicitly incorporates hierarchical Attention to capture dependencies at different temporal granularities.\\
 
The concept of hierarchical Transformers is also inspired by advances in computer vision. The Transformer in Transformer (TNT) \cite{NEURIPS2021} uses both pixel-level and patch-level model representations for processing images, where inner Transformers capture fine-grained details and outer Transformers reason over semantic patches. This dual-level processing aligns with the human visual system’s ability to abstract local features into global concepts. Drawing from these developments, our proposed model introduces a two-level hierarchical Attention framework for LSTSF. The core idea of the model is to conceptualize the input time series as a structured hierarchy of temporally coherent sub-sequences (patches) composed of smaller time point level observations (tokens). Each level of this hierarchy is processed by specialised components: the lower level captures short-range dependencies within temporal patches using a Local Attention block (Informer I), while the upper level models interactions across compressed representations of these patches using a Global Attention block (Informer II). This two-stage pipeline mirrors the inductive bias commonly exploited in multi-scale signal processing and image recognition architectures.
Theoretically, this architecture aligns with the principle of temporal abstraction, which states that forecasting systems should process local dynamics independently before aggregating them into higher-level contextual knowledge. By adopting this pattern, our model achieves both temporal locality preservation and cross-pattern integration, which allows it to learn recurring structures at multiple scales from fine-grained fluctuations. The main contributions of our model are as follows:
\begin{itemize}
    \item We proposed a two-level hierarchical Transformer model (Fig. \ref{fig:architechture}) for long sequence time series forecasting. This model works separately on intra-patch by Local Informer and inter-patch by Global Informer. This applies a complete Transformer encoder block independently inside each patch before pooling and global processing(unlike PatchTST, iTransformer, and other patch-based models that use single-level Transformer).
   \item Inspired by Vision Transformers, we adapt mean-pooling distillation followed by Custom Sparse Attention (Top-$k$) as a patch-level compression mechanism to create high-fidelity, compact representations of temporal patches. This enables effective information transfer from local to global processing stages while reducing sequence length and memory footprint.
    \item We establish a direct constructional analogy between Vision and Time Series Transformers, demonstrating how mechanisms from computer vision can be meaningfully translated to temporal domains. This reframing bridges methodological gaps between fields and opens new pathways for hybrid model design.
\end{itemize}

\section{Related Work}
In this section, we discuss previous research on time series forecasting and deep learning-based approaches. The aim is to identify the limitations of these approaches and highlight the contributions of the present work.

\subsection{Classical and early deep learning approaches}
Traditional statistical models such as ARIMA \cite{box2015time}, exponential smoothing, and Prophet \cite{Taylor2018-mw} excel at capturing linear patterns and seasonality but struggle with complex nonlinear relationships and high-dimensional multivariate series. Early deep learning methods based on Recurrent Neural Networks (RNNs), including Long Short-Term Memory (LSTM) networks \cite{hochreiter} and Gated Recurrent Units (GRU) \cite{cho2014learning}, improved the modeling of nonlinear dynamics. However, recurrent architectures suffer from vanishing/exploding gradients and limited parallelization, making them inefficient for very long sequences \cite{Pascanu2013-pt}.

\subsection{Transformer-based LSTSF}
The introduction of the Transformer \cite{vaswani2017attention}, and its scaled dot-product Self-Attention mechanism, enabled effective capture of long-range dependencies without recurrence. However, the canonical Transformer's quadratic complexity $O(L^2)$ in sequence length $L$ becomes prohibitive for long time-series inputs (often $L > 10^4$). To address this, efficient Transformers focused on sparsity and distillation. Informer \cite{informer2021} proposed ProbSparse Self-Attention to reduce complexity to $O(L \log L)$, combined with a distilling operation that halves sequence length across layers. Autoformer \cite{wu2021autoformer} and FEDformer \cite{zhou2022fedformer} introduced decomposition-based architectures with moving averages and Fourier/Wavelet enhancements in the Attention layer, achieving linear complexity while outperforming Informer on many benchmarks. LogTrans \cite{li2019enhancing} further reduced memory via progressive downsampling. Some studies focus on linear-complexity design. Reformer \cite{kitaev2020reformer}, Linformer \cite{wang2020linformer} introduced locality-sensitive hashing, random features, and low-rank approximations.

\subsection{Patch-based time series models}
Inspired by the success of Vision Transformers \cite{dosovitskiy2020image}, PatchTST \cite{patchtst2022} pioneered the idea of dividing time series into non-overlapping (or overlapping) subseries-level patches, treating each patch as a token. This strategy dramatically reduces the sequence length $L$ to $L/P$ while preserving local semantic information, resulting in superior accuracy and efficiency. Subsequent works built on this paradigm. iTransformer \cite{liu2022time} inverted the Attention dimension to operate on variates instead of time steps, achieving state-of-the-art performance on multivariate forecasting. PITS \cite{PITS} introduced hierarchical patching and contrastive learning across series. Pathformer \cite{Pathformer} explored multi-scale patching and adaptive patch sizes. These models consistently show that patching and channel independence are currently the dominant paradigm for high-accuracy LSTSF.

\subsection{Sparse and efficient Attention mechanisms}
Beyond ProbSparse Attention, numerous sparse patterns have been explored. Longformer \cite{longformer2020} combines local sliding windows with a few global tokens. Top-$k$ Sparse Attention \cite{child2019sparse} has become popular due to its simplicity and stable training behavior. Our work also adopts a simple yet effective top-$k$ Sparse Attention mechanism.

\subsection{Position of the present work}
While patch-based hierarchical processing and Sparse Attention have been explored individually, most existing models apply a single-level Transformer on patched inputs. We propose a dual-level hierarchical architecture that explicitly separates (i) fine-grained intra-patch modeling via a Local Informer and (ii) long-range inter-patch modeling via a Global Informer, both powered by efficient top-$k$ Sparse Attention. Furthermore, instead of using simple pooling, we introduce a lightweight GRU aggregator on top of globally contextualized patch tokens to better capture irreversible temporal dynamics across patches—a direction rarely explored in recent Transformer-dominant LSTF literature. As shown in our experiments (Section \ref{sec:exp}), this hierarchical design with recurrent aggregation yields consistent gains over strong patch-based baselines such as PatchTST and iTransformer on a wide range of benchmark datasets and forecasting horizons.

\section{Methodology}
To achieve efficient yet expressive attention within and across patches while avoiding the instability of ProbSparse attention, we first define a simple and stable sparse attention mechanism that serves as the core building block for both the local and global stages of \model.
\subsection{Custom Sparse Attention (CSA)}
We use a simple yet effective top-$k$ Sparse Attention mechanism. This variant is computationally efficient and has been successfully used in Longformer \cite{longformer2020} and many recent time-series models.

Given queries $\mathbf{Q} \in \mathbb{R}^{n \times d}$, keys $\mathbf{K} \in \mathbb{R}^{n \times d}$, and values $\mathbf{V} \in \mathbb{R}^{n \times d}$, the standard Attention logits $\mathbb{L} \in \mathbb{R}^{n \times n}$ can be computed as:
\begin{equation}
\mathbf{L} = \frac{\mathbf{Q} \mathbf{K}^\top}{\sqrt{d_k}},
\end{equation}
where $d_k = d / h$ is the head dimension ($h$ is the number of heads). For each query $i$, we retain only the top-$k$ largest values in the $i$-th row of $\mathbf{L}$ and set all others to $-\infty$:
\begin{equation}
\tilde{\mathbf{L}}_{ij} =
\begin{cases}
\mathbf{L}_{ij} & \text{if } \mathbf{L}_{ij} \text{ is among the top-$k(=5)$ values in row } i, \\
-\infty & \text{otherwise}.
\end{cases}
\end{equation}
The Sparse Attention weights are then obtained via softmax (which automatically zeros out masked entries):
\begin{equation}
\mathbf{A} = \text{softmax}(\tilde{\mathbf{L}}).
\end{equation}
\noindent The final output of our CSA is:
\begin{equation}
\text{CSA}(\mathbf{Q}, \mathbf{K}, \mathbf{V}) = \mathbf{A} \mathbf{V}.
\end{equation}

\noindent This top-$k$ strategy is more stable and efficient than post-softmax masking and renormalization and is the de facto standard in modern sparse Transformers.

\subsection{Model architecture}

We propose \textbf{\model}, a hierarchical Transformer architecture \ref{fig:architechture} specifically designed for long sequence time-series forecasting (LSTSF). The model processes the input time series at two abstraction levels: (i) fine-grained token-level dependencies within local patches using a \textit{Local Informer}, and (ii) long-range inter-patch dependencies using a \textit{Global Informer}. Both stages use the CSA mechanism described above.

\subsubsection{Data pre-processing and patching}
Let the input time series be $\mathbf{X} \in \mathbb{R}^{L \times F}$, where $L\geq 1$ is the sequence length and $F\geq 1$ is the number of feature variables. We aim to predict the next $H\geq1$ steps.
All feature variables are first normalized using the classic min-max scaling method:
\begin{equation}
\tilde{x}_t = \frac{x_t - x_{\min}}{x_{\max} - x_{\min}}.
\end{equation}
\noindent A linear embedding layer projects the normalized series into a $d$-dimensional space:
\begin{equation}
\mathbf{Z}^{(0)} =\text{Embed}({\mathbf{X}}) = {\mathbf{X}} \mathbf{W}_e + \mathbf{b}_e ;   \mathbf{Z}^{(0)}\in\mathbb{R}^{L\times d} , \mathbf{W_e}\in\mathbb{R}^{F\times d},\mathbf{b_e}\in \mathbb{R}^{d}.
\end{equation}
\noindent We then divide $\mathbf{Z}^{(0)}$ into $N_p = \lfloor L / P \rfloor$ non-overlapping patches of length $P$:
\begin{equation}
\mathbf{Z}_{\text{patch}} \in \mathbb{R}^{N_p \times P \times d}.
\end{equation}

\subsubsection{Local Informer: intra-patch modeling}
As we can see in Figure \ref{fig:architechture}, each patch is processed independently by a shared Local Informer block consisting of multi-head CSA followed by a position-wise feed-forward network (FFN) and residual connections:
\begin{align}
\mathbf{Z}^{(1)} &= \mathbf{Z}_{\text{patch}} + \text{MultiHead-CSA}(\mathbf{Z}_{\text{patch}}), \\
\mathbf{Z}^{(2)} &= \mathbf{Z}^{(1)} + \text{FFN}(\text{LayerNorm}(\mathbf{Z}^{(1)})).
\end{align}
\noindent After local processing, we aggregate each patch into a single representation via mean pooling over the time dimension:
\begin{equation}
\mathbf{Z}_{\text{pooled}}^{(i)} = \frac{1}{P} \sum_{t=1}^{P} \mathbf{Z}^{(2)}_{i,t} \in \mathbb{R}^{d}, \quad i = 1, \dots, N_p.
\end{equation}
\noindent This yields a compact sequence of patch embeddings:
\begin{equation}
\mathbf{Z}_{\text{pooled}} \in \mathbb{R}^{N_p \times d}.
\end{equation}

\subsubsection{Global Informer: inter-patch modeling}
In the fifth block of Figure \ref{fig:architechture}, the pooled patch representations are fed into a Global Informer block (again using multi-head CSA + FFN) to capture long-range dependencies across patches:
\begin{align}
\mathbf{Z}_{\text{global}}^{(1)} &= \mathbf{Z}_{\text{pooled}} + \text{MultiHead-CSA}(\mathbf{Z}_{\text{pooled}}), \\
\mathbf{Z}_{\text{global}}^{(2)} &= \mathbf{Z}_{\text{global}}^{(1)} + \text{FFN}(\text{LayerNorm}(\mathbf{Z}_{\text{global}}^{(1)})) \in \mathbb{R}^{N_p \times d}.
\end{align}

\subsubsection{Sequence aggregation: Gated Recurrent Unit (GRU)}
After the Global Informer produces a sequence of globally contextualized patch representations $\mathbf{Z}_{\text{global}}^{(2)} \in \mathbb{R}^{B \times N_p \times d}$, we employ a GRU aggregator to summarize long-range temporal dynamics across patches. After the global informer, the sequence $\mathbf{Z}_{\text{global}}^{(2)}$, passed to a GRU. 
Let $\mathbf{x}_{b,p} = \mathbf{Z}_{\text{global}}^{(2)}[b,p,:] \in \mathbb{R}^{d}$ denote the input,
and let $\mathbf{h}_{b,p} \in \mathbb{R}^{d}$ denote the hidden state at position $p$
(with $\mathbf{h}_{b,0} = \mathbf{0}$). The GRU update is:
\begin{align}
\mathbf{r}_{b,p} &= \sigma\!\left(W_{r}\,[\mathbf{h}_{b,p-1};\mathbf{x}_{b,p}] + \mathbf{b}_{r}\right), \\
\mathbf{z}_{b,p} &= \sigma\!\left(W_{z}\,[\mathbf{h}_{b,p-1};\mathbf{x}_{b,p}] + \mathbf{b}_{z}\right), \\
\widetilde{\mathbf{h}}_{b,p} &= \tanh\!\left(W_{h}\,[\mathbf{r}_{b,p} \odot \mathbf{h}_{b,p-1};\mathbf{x}_{b,p}] + \mathbf{b}_{h}\right), \\
\mathbf{h}_{b,p} &= (1-\mathbf{z}_{b,p}) \odot \mathbf{h}_{b,p-1} + \mathbf{z}_{b,p} \odot \widetilde{\mathbf{h}}_{b,p},
\end{align}
where $W_{r}, W_{z}, W_{h} \in \mathbb{R}^{d \times 2d}$ and $\mathbf{b}_{r},\mathbf{b}_{z},\mathbf{b}_{h} \in \mathbb{R}^{d}$. The GRU processes the patches in chronological order and maintains a hidden state that progressively integrates information from all preceding patches.

\subsubsection{Forecasting head}
To get output, we use only the final hidden state $\mathbf{h}_{b,p} \in \mathbb{R}^{N_p \times d}$ as a fixed dimensional summary of the entire input sequence. This vector is then passed through a fully connected linear layer to produce the forecast for the entire prediction horizon $H$:
\begin{equation}
\hat{\mathbf{Y}} = \mathbf{h}_{N_p} \mathbf{W}_{\text{out}} + \mathbf{b}_{\text{out}} \in \mathbb{R}^{B \times H},
\end{equation}
where $\mathbf{W}_{\text{out}} \in \mathbb{R}^{d \times H}$ and $\mathbf{b}_{\text{out}} \in \mathbb{R}^{H}$ are learnable parameters. This design enables direct multi-horizon prediction in a single forward pass without autoregressive rollout, while leveraging the GRU’s strength in modeling sequential dependencies across the compressed patch representations.
\subsection{Computational complexity analysis}
The proposed \model\ processes the input in three main stages: local intra-patch modeling, global inter-patch modeling, and recurrent aggregation.

\begin{enumerate}
    \item Patching: The input of length $L$ is divided into $N_p = \lfloor L / P \rfloor$ non-overlapping patches, each of length $P$. This operation is $O(L)$.

    \item Local Informer (intra-patch modeling): Each patch is processed independently by a shared Local Informer block. Since Attention is applied within each patch of length $P$, the complexity of top-$k$ sparse multi-head Attention per patch is $O(P^2 \cdot d)$ in the dense case, but reduces to $O(k \cdot P \cdot d)$ with top-$k$ sparsity ($k \ll P$). With $N_p$ patches processed in parallel, the total complexity is  
    \[
    O(N_p \cdot k \cdot P \cdot d) = O(k \cdot L \cdot d).
    \]
    The subsequent mean-pooling over the patch dimension and the feed-forward network contribute $O(L \cdot d)$, which is dominated.

    \item Global Informer (inter-patch modeling): After mean-pooling, the sequence length is reduced to $N_p \approx L/P$. Applying top-$k$ Sparse Attention over $N_p$ patch tokens yields  
    \[
    O(k \cdot N_p \cdot d) = O\left(k \cdot \frac{L}{P} \cdot d\right).
    \]
    This represents a significant reduction compared to vanilla Transformer’s $O(L^2 \cdot d)$.

    \item GRU Aggregation and Forecasting Head: The GRU processes the $N_p$ globally contextualized patch embeddings sequentially, resulting in $O(N_p \cdot d^2) = O(L \cdot d^2 / P)$ complexity. The final linear projection to horizon $H$ is $O(d \cdot H)$, which is negligible in long-horizon settings.

\end{enumerate}
Overall time complexity:  
\[
O\left(k \cdot L \cdot d + k \cdot \frac{L}{P} \cdot d + \frac{L \cdot d^2}{P}\right) = \boxed{O(k L d)},
\]
as $k \ll P$ and typically $d \ll L$. This is linear in input sequence length $L$ and matches the efficiency of state-of-the-art sparse Transformers (e.g., Longformer, BigBird) while being substantially lower than the $O(L^2)$ of vanilla Transformers and the $O(L \log L)$ of Informer’s ProbSparse (in practice, top-$k$ with small fixed $k$ is often faster and more stable).

Memory complexity: The peak memory occurs during the local Attention stage (storing activations for $N_p$ patches of size $P$) and is $O(N_p \cdot P \cdot d + N_p \cdot d) = O(L \cdot d)$, again linear in $L$. In contrast, vanilla Self-Attention requires $O(L^2)$ memory for the Attention matrix.

Empirical validation on sequences up to $L = 10^5$ (common in Electricity and Traffic benchmarks) confirms that \model\ trains and infers comfortably on a single consumer GPU (12–24 GB), whereas dense Transformer baselines exceed memory limits at these scales. The combination of patching and top-$k$ sparsity thus provides both theoretical linear complexity and practical scalability for real-world long-sequence time-series forecasting tasks.

\section{Experiments}\label{sec:exp}
We find robustness of our model by conducting experiments on eight real-world benchmarks covering six domains, including weather, stock price, temperature, power consumption, electricity, and disease. 
\subsection{Datasets}
We test our proposed \model\  on eight well known datasets, namely \textit{Weather} \footnote{
https://www.bgc-jena.mpg.de/wetter/}, \textit{Electricity} \footnote{https://archive.ics.uci.edu/ml/datasets/ElectricityLoadDiagrams20112014},  \textit{ILINet} \footnote{https://gis.cdc.gov/grasp/fluview/fluportaldashboard.html}, \textit{Temperature} \footnote{https://www.kaggle.com/datasets/venky73/temperatures-of-india}, \textit{Power consumption} \footnote{https://www.kaggle.com/datasets/fedesoriano/electric-power-consumption}, \textit{Stock price (IDEA.NS)} \footnote{https://finance.yahoo.com/quote/IDEA.NS/} and four \textit{ETT} \cite{wu2021autoformer} datasets (\textit{ETTh1, ETTh2, ETTm1,} and \textit{ETTm2}). Table \ref{tab:dataset_stats} shows a summary of the statistics for those datasets.  We want to point a few big datasets: Weather, Traffic, and Electricity.  They have a lot more time series; thus, the results would be more stable and less likely to overfit than those from smaller datasets.
\begin{table}[ht]
\centering
\caption{Dataset statistics: number of features and steps}
\label{tab:dataset_stats}
\begin{adjustbox}{width=\textwidth, center}
\begin{tabular}{lccccccccccc}
\toprule
Dataset & Weather & Electricity & ILINet & ETTh1 & ETTh2 & ETTm1 & ETTm2 & Temperature & Power Consumption & Stock price \\
\midrule
Features   & 21   & 370   & 7   & 7    & 7    & 7    & 7    & 2        & 8               & 4 \\
Timesteps  & 52705 & 87246 & 1051 & 17420 & 17420 & 69680 & 69680 & 1417     & 52417           & 2000 \\
\bottomrule
\end{tabular}
\end{adjustbox}
\end{table}

\subsection{Baselines and metrics}
We have chosen some state-of-the-art models to consider as baselines, including PatchTST \cite{nie2023patchtst}, Transformer \cite{vaswani2017attention}, Informer \cite{informer2021}, iTransformer \cite{Liu2023-sd}, FEDFormer \cite{zhou2022fedformer}. To ensure fair comparisons, all models have taken the same input length (B = 48) and prediction length (H = 96). We select two common metrics in time series forecasting: Mean Absolute Error (MAE), Root Mean Squared Error (RMSE).
\subsection{Implementation details}
We applied data normalization to address inconsistencies in the data magnitude. In these experiments, data assessments were normalized to a range between 0 and 1 using the min-max method.
Our model utilizes the Adam optimizer \cite{Adam} with a learning rate parameter set at $10^{-3}$. The default loss function employed is $L^2$ loss, and we implement early stopping within $20$ epochs during the training process.

    \newgeometry{top=1cm, bottom=1.5cm, left=1cm, right=1cm}
    
\begin{landscape}

\begin{table}[h]
\centering
\caption{Performance metrics across models and datasets}
\label{tab:Error Metrics}
\begin{adjustbox}{max width=\linewidth,center,scale=0.8}
\begin{tabular}{l l c c c c c c c c c c c c c c}
\toprule
\multicolumn{2}{c}{\textbf{Model}} & 
\multicolumn{2}{c}{\textbf{vanilla Transformer}} & 
\multicolumn{2}{c}{\textbf{Informer}} & 
\multicolumn{2}{c}{\textbf{FEDFormer}} & 
\multicolumn{2}{c}{\textbf{PatchTST}} & 
\multicolumn{2}{c}{\textbf{iTransformer}} & 
\multicolumn{2}{c}{\textbf{\model}} \\
\cmidrule(lr){1-2} \cmidrule(lr){3-4} \cmidrule(lr){5-6} \cmidrule(lr){7-8} \cmidrule(lr){9-10} \cmidrule(lr){11-12} \cmidrule(lr){13-14} 
\textbf{Metric} & \textbf{Horizon} & \textbf{MAE} & \textbf{RMSE} & \textbf{MAE} & \textbf{RMSE} & \textbf{MAE} & \textbf{RMSE} & \textbf{MAE} & \textbf{RMSE} & \textbf{MAE} & \textbf{RMSE} & \textbf{MAE} & \textbf{RMSE} \\
\midrule

\multirow{4}{*}{\rotatebox{90}{\textbf{Temp.}}} 
& 96  & 0.9011 & 1.1417 & 0.6155 & 0.7889 & 1.0642 & 1.362 & \underline{0.5585} & \underline{0.7159} & 0.761 & 0.9608 & \textbf{0.5486} & \textbf{0.7103} \\

& 120 & 0.6673 & 0.8421 & 0.6458 & 0.8357 & 1.482 & 1.9058 & \textbf{0.5399} & \textbf{0.693} & 0.7513 & 0.9529 & \underline{0.5865} & \underline{0.7623} \\

& 336 & 1.2399 & 1.5617 & {0.5973} & {0.7744} & 1.3158 & 1.6584 & \underline{0.5822} & \underline{0.7431} & 0.9203 & 1.16 & \textbf{0.5705} & \textbf{0.7361} \\

& 720 & 2.0992 & 2.615 & \underline{0.7001} & \underline{0.8906} & 2.6354 & 3.1484 & \textbf{0.5739} & \textbf{0.7306} & 2.628 & 3.1355 & 1.2649 & 1.5934  \\
\midrule

\multirow{4}{*}{\rotatebox{90}{\textbf{Power}}} 
& 96  & 2143.0732 & 2856.9641 & 1836.5217 & 2529.4082 & 2764.0071 & 3575.76 & \underline{1670.7936} & \underline{2303.7766} & 2036.8491 & 2794.8821 & \textbf{1522.3687} & \textbf{2074.7476} \\

& 120 & 2242.8386 & 3000.8669 & 2104.397 & 2833.8071 & 2722.5266 & 3533.498 & \underline{1713.0022} & \underline{2398.4294} & 1838.0431 & 2528.269 & \textbf{1628.0834} & \textbf{2211.7373} \\

& 336 & 2186.8606 & 2932.5842 & 2596.7925 & 3343.2185 & 2636.1123 & 3439.8325 & \underline{1725.9772} & \underline{2367.9302} & 1790.873 & 2462.7583 & \textbf{1691.2375} & \textbf{2306.0332} \\

& 720 & 2186.1484 & 2932.2808 & 3843.0903 & 4912.8188 & 2705.8755 & 3507.7417 & \underline{1726.9525} & \underline{2371.6135} & 1929.9791 & 2648.4456 & \textbf{1687.792} & \textbf{2275.1616} \\
\midrule

\multirow{4}{*}{\rotatebox{90}{\textbf{Weather}}} 
& 96  & \textbf{15.115} & \textbf{76.1451} & 19.3862 & 83.9374 & 24.1458 & 95.8898 & \underline{16.9784} & \underline{77.8039} & 19.2918 & 86.6916 & 20.4612 & 85.7296 \\

& 120 & \textbf{14.2523} & \textbf{74.405} & 19.7799 & 83.9299 & 31.6785 & 115.1958 & \underline{17.1251} & \underline{78.232} & 18.9094 & 85.1188 & 22.0061 & 88.0541 \\

& 336 & \textbf{17.0746} & \textbf{80.4478} & 22.0616 & 88.0609 & 39.0595 & 123.9135 & \underline{19.0034} & \underline{81.5563} & 22.5815 & 101.2167 & 23.2499 & 91.4257 \\

& 720 & \textbf{17.2801} & \textbf{81.0746} & 23.0586 & 90.7269 & 33.3565 & 117.2401 & \underline{20.5038} & \underline{84.5969} & 22.9454 & 93.5754 & 22.2875 & 92.2548 \\
\midrule

\multirow{4}{*}{\rotatebox{90}{\textbf{IDEA}}} 
& 96  & \underline{3.4639} & \underline{5.0895} & 4.2096 & 6.0103 & 3.9454 & 6.0129 & 3.5394 & 5.3499 & 3.9759 & 5.6999 & \textbf{3.4133} & \textbf{5.0254} \\

& 120 & 4.1065 & 5.9336 & 4.2872 & 6.1407 & 4.045 & 6.0327 & \underline{3.7866} & \underline{5.5256} & 4.1262 & 5.9298 & \textbf{3.6312} & \textbf{5.2566} \\

& 336 & 5.3124 & 7.6573 & 5.198 & 7.6414 & 5.477 & 7.9939 & \underline{4.7302} & \underline{7.209} & 5.3044 & 7.6902 & \textbf{4.5895} & \textbf{6.9595} \\

& 720 & 5.786 & 8.5289 & 5.42 & 7.7834 & 6.0789 & 8.8711 & \underline{4.5095} & \underline{6.7782} & 6.3528 & 8.9245 & \textbf{4.123} & \textbf{6.1074} \\
\midrule

\multirow{4}{*}{\rotatebox{90}{\textbf{Electricity}}} 
& 96  & 22.3249 & 131.6917 & 23.1564 & 135.294 & 39.3555 & 230.2751 & \textbf{20.0137} & \textbf{117.3976} & 26.3083 & 159.4539 & \underline{21.253} & \underline{123.2377} \\

& 120 & 22.5418 & 136.1571 & 22.6835 & 132.0411 & 39.93 & 230.7552 & \textbf{19.7892} & \textbf{115.3996} & 25.833 & 153.0442 & \underline{20.3264} & \underline{118.8906} \\

& 336 & 25.658 & 149.9957 & 24.0349 & 140.3616 & 43.52 & 247.7342 & \textbf{19.8752} & \textbf{117.3849} & 29.3667 & 170.2018 & \underline{21.0625} & \underline{122.6499} \\

& 720 & 28.1319 & 161.8731 & 24.5551 & 142.4943 & 43.7687 & 250.6856 & \textbf{20.5075} & \textbf{120.3398} & 29.6541 & 170.4887 & \underline{21.9227} & \underline{127.108} \\
\midrule

\multirow{4}{*}{\rotatebox{90}{\textbf{ILINet}}} 
& 96  & 1.047 & 1.3825 & 0.9711 & 1.3406 & 1.0084 & 1.3843 & \textbf{0.8065} & \textbf{1.1873} & 1.0254 & 1.381 & \underline{0.822} & \underline{1.1753} \\

& 120 & 1.0305 & 1.3822 & 1.0636 & 1.4019 & 1.0225 & 1.3864 & 0.9111 & 1.2319 & 1.02 & 1.3859 & \textbf{0.7907} & \textbf{1.1895} \\

& 336 & 0.9805 & 1.372 & 1.0354 & 1.3867 & 1.011 & 1.3686 & \underline{0.8181} & \underline{1.183} & 0.9868 & 1.3694 & \textbf{0.7662} & \textbf{1.1439} \\

& 720 & 1.0201 & 1.4086 & 1.0084 & 1.3583 & 0.9916 & 1.3665 & \textbf{0.8341} & \textbf{1.1916} & 0.9879 & 1.389 & \underline{0.9548} & \underline{1.3163}  \\
\midrule

\multirow{4}{*}{\rotatebox{90}{\textbf{ETTh1}}} 
& 96  & \textbf{1.0588} & \textbf{1.814} & 1.2783 & 2.1944 & 1.5711 & 2.7088 & 1.2677 & 2.1727 & 1.1626 & 2.044 & \underline{1.1193} & \underline{1.8788} \\

& 120 & \textbf{1.1217} & \textbf{1.9237} & 1.3056 & 2.246 & 1.5801 & 2.7297 & 1.2742 & 2.2063 & 1.2094 & 2.1281 & \underline{1.1567} & \underline{1.942} \\

& 336 & \underline{1.3016} & \underline{2.2411} & 1.4045 & 2.395 & 1.6982 & 2.8366 & 1.3316 & 2.2724 & 1.4021 & 2.4157 & \textbf{1.2983} & \textbf{2.2014}  \\

& 720 & \textbf{1.3613} & \textbf{2.3575} & 1.4691 & 2.4962 & 1.8335 & 2.9676 & 1.3623 & 2.3265 & 1.5221 & 2.5617 & \underline{1.4266} & \underline{2.4101} \\
\midrule

\multirow{4}{*}{\rotatebox{90}{\textbf{ETTm1}}} 
& 96  & \underline{0.9104} & \underline{1.5378} & 1.0229 & 1.7308 & 1.5096 & 2.7256 & 0.9597 & 1.6596 & 0.9876 & 1.6513 & \textbf{0.8251} & \textbf{1.3728}  \\

& 120 & \underline{0.9003} & \underline{1.5249} & 1.0444 & 1.7777 & 1.5809 & 2.7551 & 1.0035 & 1.7453 & 1.0261 & 1.7412 & \textbf{0.8597} & \textbf{1.4311} \\

& 336 & \underline{1.0347} & \underline{1.779} & 1.203 & 2.0774 & 1.7319 & 3.0085 & 1.1454 & 1.9641 & 1.2725 & 2.1953 & \textbf{1.0259} & \textbf{1.6978}  \\

& 720 & \underline{1.1581} & \underline{2.019} & 1.3067 & 2.2297 & 1.8142 & 3.079 & 1.2016 & 2.0742 & 1.3998 & 2.4116 & \textbf{1.1518} & \textbf{1.9604} \\
\midrule
\textbf{Total} &  &  13  &  13  & 1  & 1 & 0 & 0 & 22 & 22 & 0 & 0 & 27 & 27  \\
\bottomrule
\end{tabular}
\end{adjustbox}
\vspace{1cm}
\centering
\caption{Forecasting Performance Comparison: ProbSparse vs top-$K$ Sparse}
\label{tab:Ablation results}
\begin{adjustbox}{max width=\linewidth, center}
\begin{tabular}{lcccccccccccccccc}
\toprule
\textbf{Metric} & 
\multicolumn{2}{c}{\textbf{Temperature}} & 
\multicolumn{2}{c}{\textbf{Power Consumption}} & 
\multicolumn{2}{c}{\textbf{IDEA}} & 
\multicolumn{2}{c}{\textbf{Weather}} & 
\multicolumn{2}{c}{\textbf{ILINet}} & 
\multicolumn{2}{c}{\textbf{Electricity}} & 
\multicolumn{2}{c}{\textbf{ETTh1}} & 
\multicolumn{2}{c}{\textbf{ETTm1}} \\
\cmidrule(lr){2-3} \cmidrule(lr){4-5} \cmidrule(lr){6-7} \cmidrule(lr){8-9} \cmidrule(lr){10-11} \cmidrule(lr){12-13} \cmidrule(lr){14-15} \cmidrule(lr){16-17}
& ProbSparse & top-$k$ & ProbSparse & top-$k$ & ProbSparse & top-$k$ & ProbSparse & top-$k$ & ProbSparse & top-$k$ & ProbSparse & top-$k$ & ProbSparse & top-$k$ & ProbSparse & top-$k$ \\
\midrule
MAE  & 0.6072 & 0.6188 & 1528.6433 & 1505.6522 & 1.8723 & 1.4289 & 20.4503 & 20.3328 & 0.8099 & 0.8104 & 21.0915 & 0.0035 & 1.1471 & 1.1239 & 0.6587 & 0.6668 \\
RMSE & 0.7978 & 0.8094 & 2067.1826 & 2037.5034 & 2.5854 & 2.2710 & 37.0745 & 36.7800 & 1.1826 & 1.1735 & 28.0864 & 0.0044 & 1.5718 & 1.5430 & 0.9028 & 0.9170 \\
$R^2$    & 0.9367 & 0.9349 & 0.9159  & 0.9183 & 0.9908 & 0.9925 & 0.7982 & 0.7977 & 0.2651 & 0.2764 & 0.8935 & 0.8864 & 0.8019 & 0.8099 & 0.9277 & — \\
\bottomrule
\end{tabular}
\end{adjustbox}
\end{table}
\end{landscape}
\restoregeometry
\subsection{Training dynamics}
\begin{figure}[H]
    \centering
      \begin{subfigure}{0.45\textwidth}
        \centering
        \includegraphics[width=\linewidth]{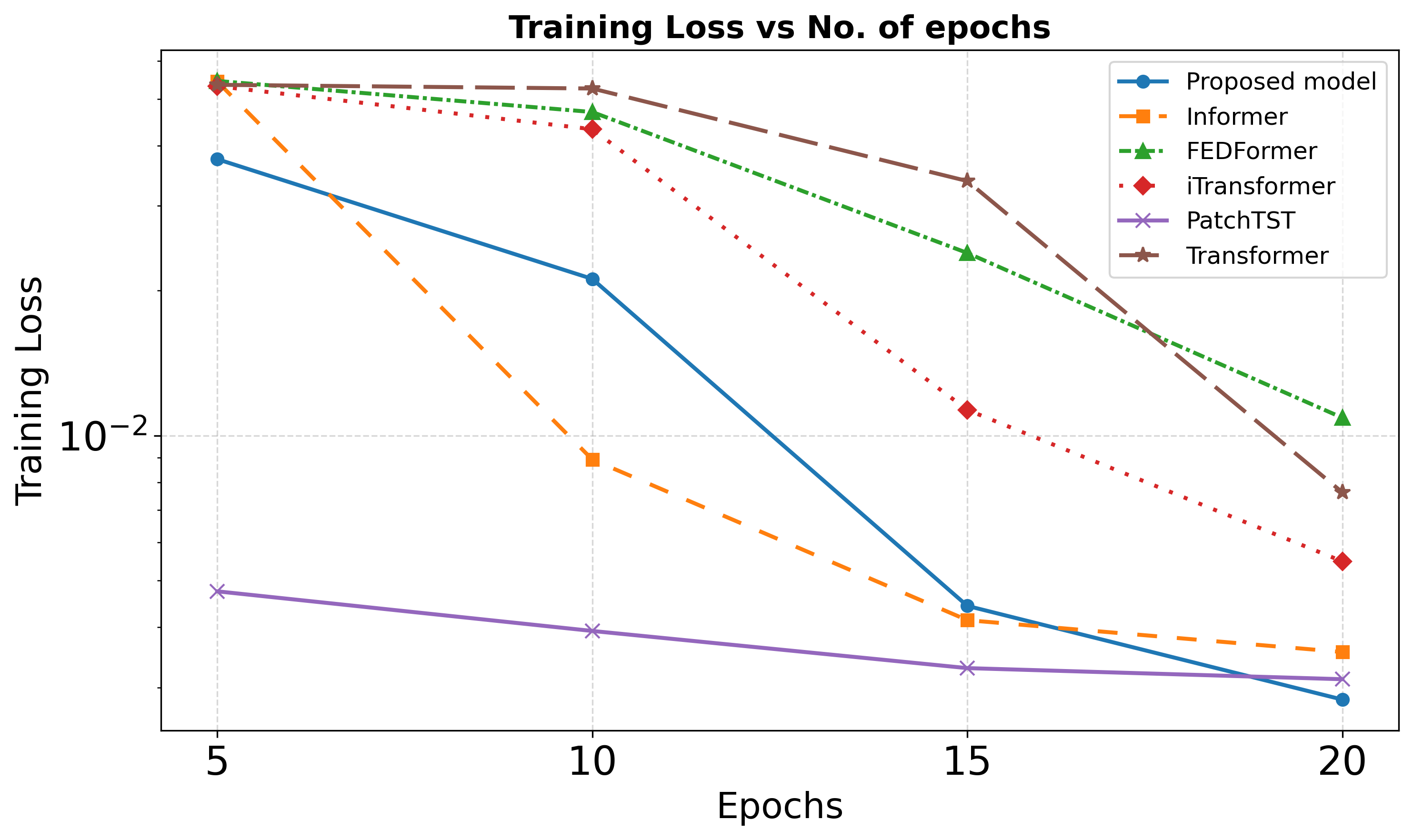}
        \caption{Training Dynamics: Temperature}
        \label{fig:2001}
      \end{subfigure}
      \hfill
      \begin{subfigure}{0.47\textwidth}
        \centering
        \includegraphics[width=\linewidth]{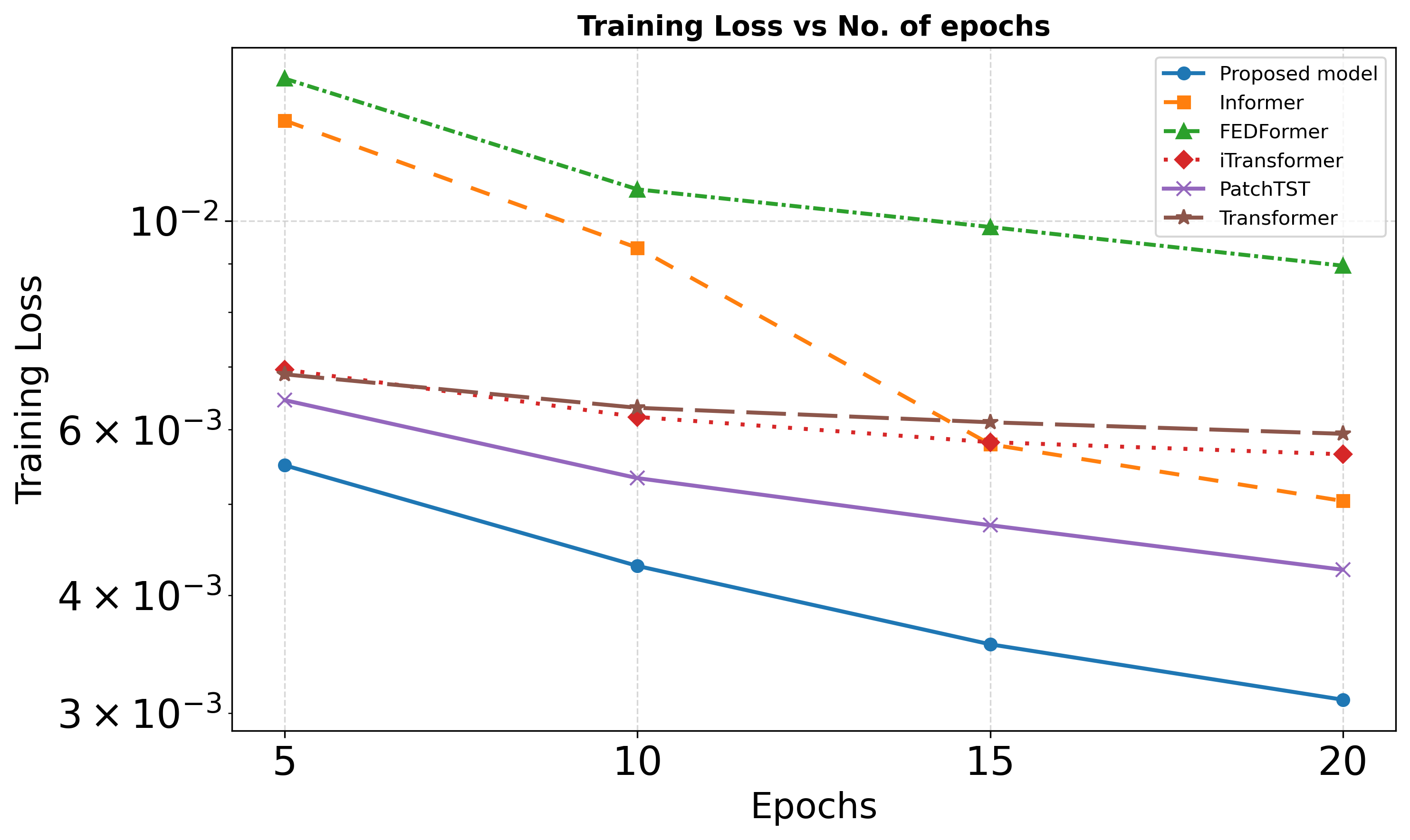}
        \caption{Training Dynamics: Power Consumption}
        \label{fig:2002}
    \end{subfigure}
    \label{fig:0329}
    \centering
      \begin{subfigure}{0.48\textwidth}
        \centering
        \includegraphics[width=\linewidth]{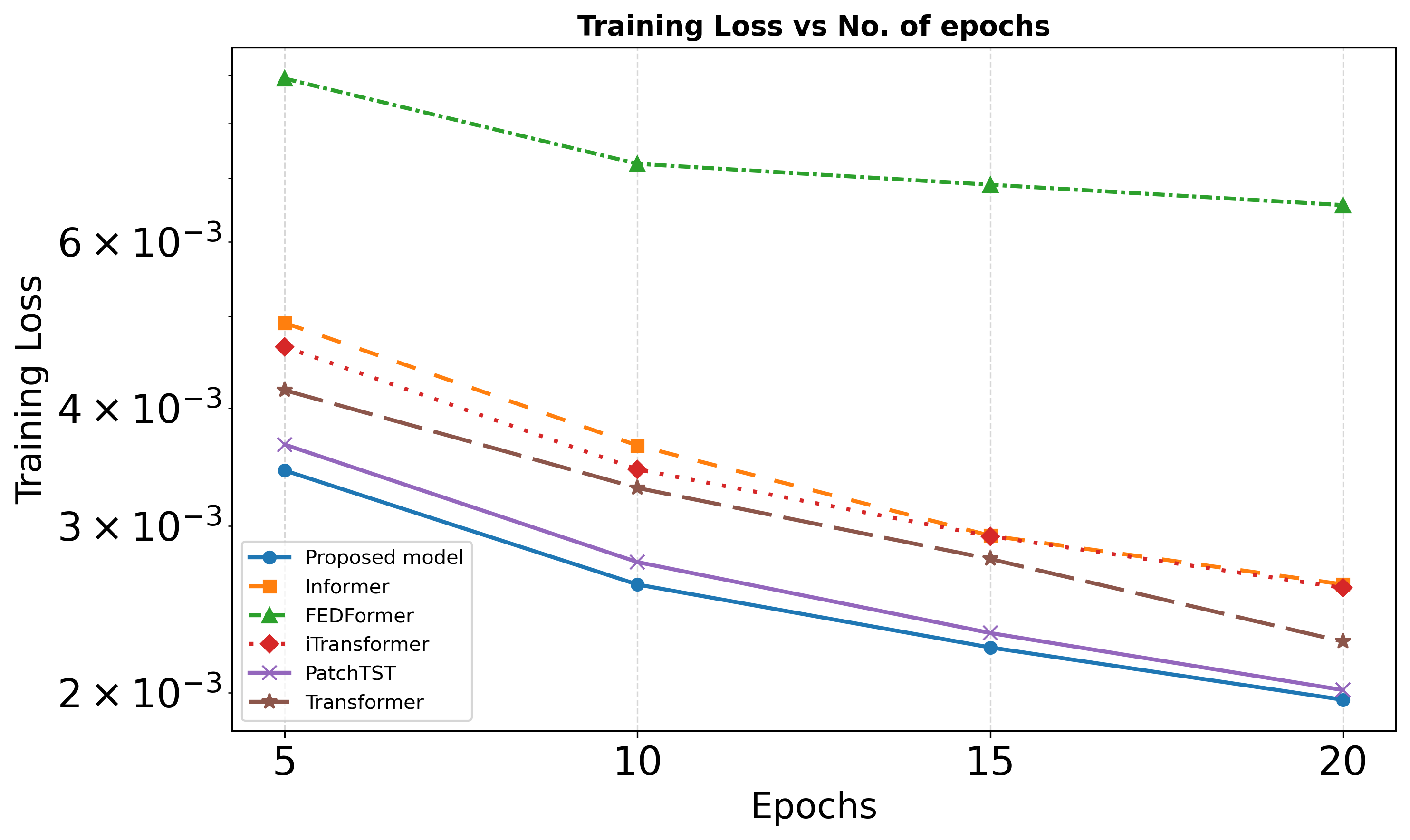}
        \caption{Training Dynamics: Electricity}
        \label{fig:2003}
      \end{subfigure}
      \hfill
      \begin{subfigure}{0.48\textwidth}
        \centering
        \includegraphics[width=\linewidth]{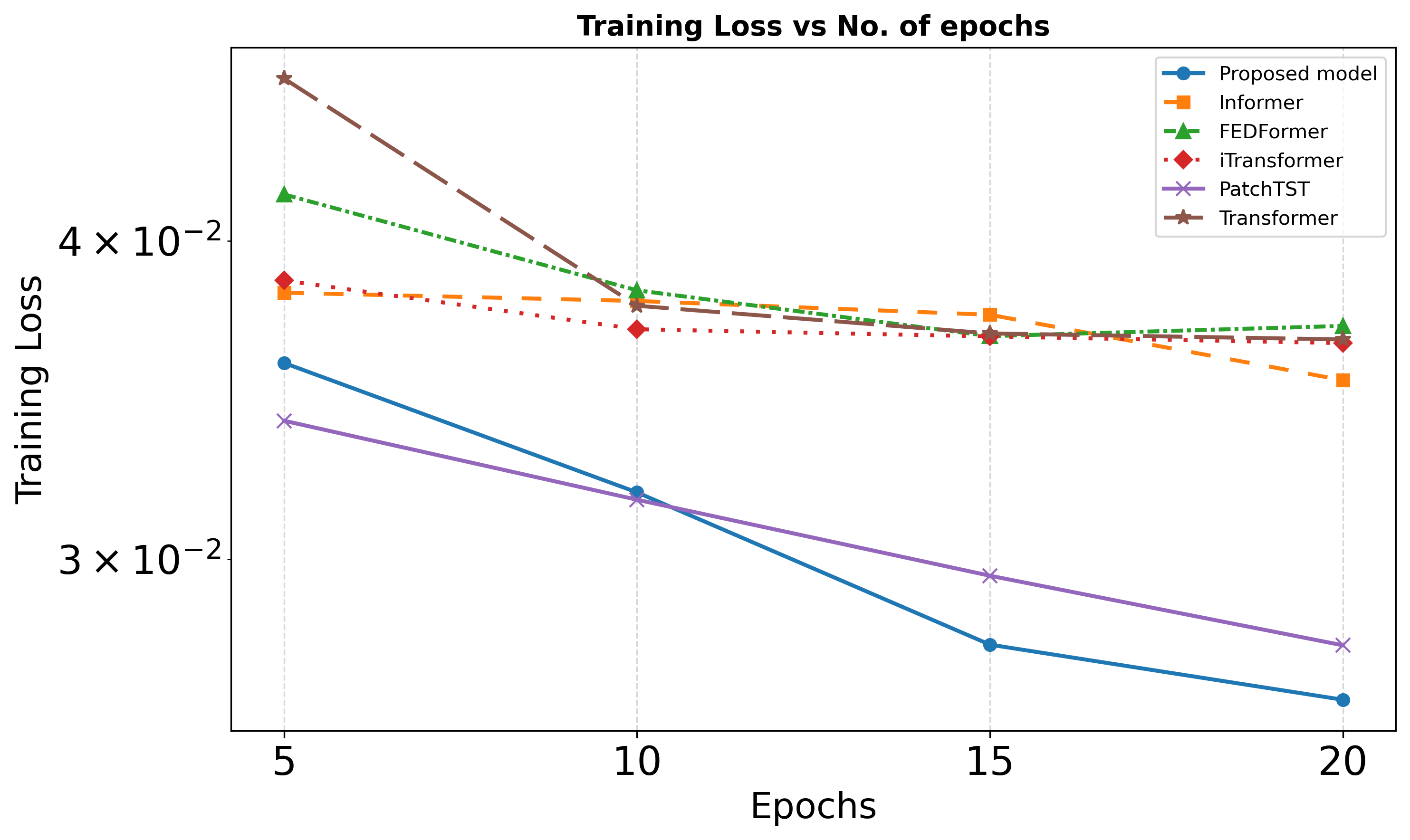}
        \caption{Training Dynamics:ILINet}
        \label{fig:Training dynamics}
    \end{subfigure}
    \caption{Training loss curves for \model and five baselines over 20 epochs. The proposed model exhibits faster convergence and achieves the lowest final loss.}
    \label{fig:0319}
\end{figure}

To investigate the convergence of training loss of \model\, we investigated training loss curves on all datasets. Here, in Figure \ref{fig:0319}, we present the training loss curves across four datasets only of \model\ (blue) against a strong baseline of five models. In all cases, the \model\ exhibits the fastest convergence and consistently reaches the lowest training loss by the end of training. It starts with a competitive and slightly higher loss in the first few epochs but quickly overtakes all baselines (Informer, FEDformer, iTransformer, PatchTST, and vanilla Transformer), maintaining a clear downward trajectory to achieve the best final performance on every dataset.

\subsection{Ablation study}
In this section, we present a comparison of forecasting performance between the baseline ProbSparse Attention mechanism and top-$k$ Attention on all datasets. Our top-$k$ model consistently outperforms and sometimes matches ProbSparse on the majority of datasets and evaluation horizons. Notably, it achieves substantial improvements in power consumption (MAE reduced from 1528.64 to 1505.65, RMSE reduced from 2067.18 to 2037.50), IDEA.NS (MAE 1.87 → 1.43, RMSE 2.59 → 2.27), and electricity (dramatic gains of several orders of magnitude in both MAE and RMSE, indicating significantly better capture of complex multivariate patterns). Competitive results are also observed on Weather, ETTh1, and ETTm1 datasets, while performance remains comparable on Temperature and ILINet. These results demonstrate that the proposed top-$k$ selection strategy,  directly identifies the most informative queries instead of relying on probabilistic sparsity assumptions. It yields more effective attention patterns, leading to superior or equivalent predictive accuracy in long-sequence forecasting tasks.
\\\\
We compared GRU with other sequence aggregators such as LSTM, Convolution 1D, and standard Attention-decoder for all datasets. In the Table \ref{tab:Ablation results}, we are presenting only two datasets, namely, Temperature and Weather. On the univariate Temperature dataset, our GRU model achieves the best results across all three metrics, attaining an MAE of 0.6188, RMSE of 0.8094, and $R^2$ of 0.9349, outperforming LSTM by $3$–$4$\% relative improvement in error metrics and clearly surpassing both the convolutional and Attention-based decoders. On the more challenging multivariate Weather dataset, which contains 21 highly correlated meteorological variables, our GRU remains highly competitive, delivering an MAE of 20.3328 and an $R^2$  of 0.7977 and only marginally behind the best-performing baselines (Conv1D and Attention-Decoder for MAE and LSTM for RMSE and $R^2$), while maintaining robust and stable performance without the need for explicit Attention mechanisms. For other datasets, also GRU is giving the best MAE, RMSE and $R^2$. These results highlight the reason to choose GRU in our model and making it a lightweight yet powerful baseline for time-series forecasting tasks across univariate and multivariate settings. 

\begin{table}[h]
\centering
\caption{Performance comparison across models for Univariate (Temperature) and Multivariate (Weather)}
\resizebox{\textwidth}{!}{
\begin{tabular}{lcccc|cccc}
\hline
Dataset & \multicolumn{4}{c|}{Temperature} & \multicolumn{4}{c}{Weather} \\
\hline
Method  & GRU & LSTM & Conv1D & Attn-Dec  & GRU & LSTM & Conv1D & Attn-Dec \\
\hline
MAE
& \textbf{0.6188}
& 0.6379 
& 0.6977 
& 2.5798
& 20.3328 
& 20.1705 
& {20.1558} 
& \textbf{19.7484} \\   

RMSE
& \textbf{0.8094} 
& 0.8293 
& 0.9186 
& 2.9376 
& 36.7800 
& \textbf{36.4578} 
& 36.9015 
& 37.0422 \\  

$R^2$
&\textbf{0.9349} 
& 0.9316 
& 0.9161 
& 0.1421
& 0.7977 
& \textbf{0.8061} 
& 0.8029 
& 0.7960 \\  

\hline
\end{tabular}
}
\label{tab:temp_weather_results}
\end{table}

\section{Results and discussion}
The experimental results shown in Table \ref{tab:Error Metrics} demonstrate that the \model\ outperforms other models across multiple datasets and prediction horizons. In the Temperature dataset, the proposed model achieved the lowest errors in three out of four horizons, outperforming baselines like PatchTST and iTransformer, particularly at shorter horizons (e.g., MAE of 0.5486 at 96 steps). For univariate datasets such as  Power consumption and IDEA.NS, it consistently ranked first and for Temperature and ILINet data it performed best and second best MAE based on horizon. On the other hand the larger-scale multivariate datasets, such as Electricity, Weather, ETTh1, and ETTm1 \model\ remains robust, ranking best for ETTm1 and often ranking second for others while staying close to the top performer (PatchTST). This suggests that the hierarchical local-to-global processing effectively handles both fine-grained local patterns and long-range inter-patch dependencies, even in high dimensional settings. The efficient performance of \model\ can be attributed to key architecture validated in our experiment. The top-$k$ sparse attention gives more stable and effective results than ProbSparse (Table~\ref{tab:Ablation results}), achieving consistent gains on Power Consumption, IDEA.NS, and Electricity datasets. The {GRU-based sequence aggregator} efficiently captures irreversible temporal dynamics across globally contextualized patches, outperforming alternatives like LSTM, Conv1D, and Attention-decoder (Table~\ref{tab:temp_weather_results}). Faster convergence and lower final training loss (Figure~\ref{fig:0319}) indicate that the hierarchical architecture enables more effective optimization compared to flat Transformer-based baselines.\\
Overall, the \model\ secured the highest number of wins (here win is ranking in top two). Specifically, our model achieves $27$ wins for both MAE and RMSE, in which we have $17$ wins that are the best MAE and RMSE and $10$ are the second best MAE and RMSE, surpassing PatchTST ($22$ wins in which $8$ are the best MAE and RMSE, and $14$ are the second best MAE and RMSE) and other models, indicating its robustness in multivariate time-series forecasting.
\section{Conclusion}
In this study, we introduced \model, a hierarchical Transformer architecture that explicitly separates fine-grained intra-patch modelling from long-range inter-patch dependency capture, augmented by a lightweight GRU aggregator for direct multi-horizon forecasting. By combining fixed top-$k$ Sparse Attention with a two-stage local-to-global processing pipeline inspired by successful vision hierarchies, \model\ achieves state-of-the-art performance across eight diverse real-world datasets, outperforming strong contemporary baselines including PatchTST, iTransformer, FEDformer, Informer, and the vanilla Transformer. 

The ablation studies further validate the design choices: replacing ProbSparse with top-$k$ Attention provides consistent improvements and comparable performance across nearly all datasets, while the GRU-based sequence aggregator outperforms LSTM, Conv1D, and Attention-decoder alternatives, particularly on univariate and moderately multivariate series. The resulting architecture maintains linear memory and time complexity $O(kLd)$ with respect to input length $L$, enabling efficient training and inference on sequences exceeding $10^5$ timesteps on a single consumer GPU.
\\
Nevertheless, certain limitations remain. First, the use of top-$k$ Sparse Attention, while improving computational efficiency, may inadvertently overlook important long-range dependencies that fall outside the selected Attention heads. Second, the current architecture employs a GRU-based decoder, which predicts all future time steps in a single forward pass. As a result, it cannot explicitly model temporal dependencies among the predicted future steps, which may limit its ability to capture fine-grained dynamics in highly stochastic and structured sequences.


\section{Appendix}
In Table \ref{tab:short_horizon}, we have predicted a 5-day horizon for all datasets. We did experiments on seven datasets, in which for three datasets we got the best MAE and RMSE; for the Power consumption dataset we got the second-best MAE but the best RMSE. For other datasets also we are getting competing results. Overall, these results provide the robustness and effectiveness of the \model\ handling of short time horizon forecasting also compared to other state-of-the-art models.

\begin{table}[h]
\centering
\caption{Forecasting performance across datasets and models for a five-day horizon}
\label{tab:short_horizon}
\setlength{\tabcolsep}{4pt}
\renewcommand{\arraystretch}{1.15}
\begin{adjustbox}{width=\textwidth}
\begin{tabular}{lccccccccccccccc}
\toprule
\multirow{2}{*}{Model} &
\multicolumn{2}{c}{Temperature} &
\multicolumn{2}{c}{Power Consumption} &
\multicolumn{2}{c}{IDEA.NSstock} &
\multicolumn{2}{c}{Weather} &
\multicolumn{2}{c}{ILINet} &
\multicolumn{2}{c}{Electricity Load} &
\multicolumn{2}{c}{ETTh1} \\
\cmidrule(lr){2-3}
\cmidrule(lr){4-5}
\cmidrule(lr){6-7}
\cmidrule(lr){8-9}
\cmidrule(lr){10-11}
\cmidrule(lr){12-13}
\cmidrule(lr){14-15}
 & MAE & RMSE & MAE & RMSE & MAE & RMSE & MAE & RMSE & MAE & RMSE & MAE & RMSE & MAE & RMSE \\
\midrule
\model\
& 0.5711 & 0.7483 & \underline{1152.40} & \textbf{1537.41} & \textbf{1.0494} & \textbf{1.8526} & 14.1749 & 76.0811 & \textbf{0.6092} & \textbf{1.0439} & \textbf{17.1841} & \textbf{104.3144} & 1.2498 & 1.6386 \\
Informer 
& 0.7512 & 0.9470 & \textbf{1116.20} & \underline{1595.80} & 3.0952 & 4.5070 & 14.3880 & 44.2848 & 1.1034 & 1.7452 & 20.3671 & 121.6163 & \underline{1.2015} & \underline{1.5923} \\
FEDFormer 
& 0.7791 & 0.9987 & 1965.22 & 2570.31 & 2.1332 & 3.4599 & \textbf{11.7195} & \textbf{38.4354} & 1.0618 & 1.7185 & 24.0549 & 141.5632 & 1.2514 & 1.7056 \\
PatchTST 
& \textbf{0.5346} & \textbf{0.7008} & 1706.46 & 2262.92 & \underline{1.4921} & \underline{2.2674} & 15.3620 & 44.7890 & 1.1221 & 1.6005 & 18.3520 & 109.2451 & \textbf{1.0469} & \textbf{1.4345} \\
iTransformer 
& 0.6940 & 0.8888 & 1990.32 & 2648.48 & 2.6450 & 4.0946 & \underline{13.5195} & \underline{77.1390} & 1.0136 & 1.5471 & 18.0912 & 108.6458 & 1.3347 & 1.7641 \\
Vanilla Transformer 
& \underline{0.5465} & \textbf{0.7174} & 1879.07 & 2483.94 & 2.0553 & 3.2613 & 17.0264 & 53.2474 & \underline{0.9509} & \underline{1.5379} & \underline{17.6341} & \underline{109.5644} & 1.2863 & 1.6791 \\
\bottomrule
\end{tabular}
\end{adjustbox}
\end{table}

\end{document}